%% file: paper_template.tex
\documentclass[conference]{IEEEtran}
\input{defs}
\usepackage{times}
\usepackage{amsfonts}
\usepackage{amsmath}
\usepackage[numbers]{natbib}
\usepackage{multicol}
\usepackage[bookmarks=true]{hyperref}

\hypersetup{draft}

\begin{document}

\title{Iterative Reinforcement Learning Based Design of 
\\Dynamic Locomotion Skills for Cassie}
\author{Author Names Omitted for Anonymous Review. Paper-ID 46}

\author{\authorblockN{Zhaoming Xie\authorrefmark{1},
Patrick Clary\authorrefmark{2},
Jeremy Dao\authorrefmark{2}, 
Pedro Morais\authorrefmark{2}, Jonathan Hurst\authorrefmark{2} and
Michiel van de Panne\authorrefmark{1}}
\authorblockA{\authorrefmark{1}Department of Computer Science\\
University of British Columbia,
Vancouver, BC, Canda\\ Email: \{zxie47, van\}@cs.ubc.ca}
\authorblockA{\authorrefmark{2}Dynamic Robotics Laboratory\\Oregon State University, Corvallis, Oregon, USA\\
Email: \{claryp,daoje,autranep,jonathan.hurst\}@oregonstate.edu}}



%

\maketitle

\input{abstract}

\IEEEpeerreviewmaketitle

\input{introduction}

\input{related}

\input{preliminaries}

\input{methods}

\input{set_up}

\input{compress_distill}

\input{rl_il}





\input{conclusions}

\section*{Acknowledgments}


\bibliographystyle{plainnat}
\bibliography{references}

\end{document}

%% file: defs.tex
\usepackage[usenames] {color}
\usepackage{verbatim}
\usepackage{siunitx} 

\definecolor{purple}{rgb}{0.4,0.2,0.8}

\newcommand{\norm}[1]{\left\lVert#1\right\rVert}

\long\def\ignorethis#1{}

\usepackage{algorithm}
\usepackage[noend]{algpseudocode}
\usepackage{graphicx}
\usepackage{subcaption}

\makeatletter
\def\BState{\State\hskip-\ALG@thistlm}
\makeatother

\newcommand{\pctab}{\hspace{0.2in}}



%% file: abstract.tex
\begin{abstract}
Deep reinforcement learning (DRL) is a promising approach for developing legged locomotion skills. 
However, the iterative design process that is inevitable in practice is poorly supported
by the default methodology. It is difficult to predict the outcomes of changes made to the 
reward functions, policy architectures, and the set of tasks being trained on. 
In this paper, we propose a practical method that allows the reward function to be fully redefined
on each successive design iteration while limiting the deviation from the previous iteration.
We characterize policies via sets of Deterministic Action Stochastic State (DASS) tuples,
which represent the deterministic policy state-action pairs as sampled from the states
visited by the trained stochastic policy. New policies are trained using a policy gradient algorithm
which then mixes RL-based policy gradients with gradient updates defined by the DASS tuples. 
The tuples also allow for robust policy distillation to new network architectures.
We demonstrate the effectiveness of this iterative-design approach on the bipedal robot Cassie, 
achieving stable walking with different gait styles at various speeds.
We demonstrate the successful transfer of policies learned in simulation to the 
physical robot without any dynamics randomization, and that variable-speed walking policies
for the physical robot can be represented by a small dataset of 5-10k tuples. 

\end{abstract}

%% file: introduction.tex
\section{Introduction}
Recent success in deep reinforcement learning (DRL) has inspired much work towards constructing 
locomotion policies for legged robots. Impressive results have been demonstrated on 
planar bipeds~\cite{2018-arxiv-rl_on_atrias}, 
quadruped robots, \cite{2018-rss-sim_to_real,2019-scirobotics-anymal_learning}, 
and 6-legged robots~\cite{2019-iclr-learning_to_adapt}. 
However, these systems are relatively stable in comparison to human-scale bipeds,
for which convincing demonstrations of DRL methods to
dynamic locomotion on real hardware are still lacking, to the best of our knowledge.
Nevertheless, a variety of results in simulation point to the promise of DRL methods
in this area, e.g., ~\cite{2018-SIGGRAPH-symmetry,2017-TOG-deepLoco,2018-IROS-cassie,2017-arxiv-parkour}.
In practice, a multitude of issues can preclude the successful transfer of policies from simulation
to the physical robot, including state estimation, modeling discrepancies, and motions 
that can cause excessive wear on hardware.

\begin{figure}[t]%
\centering
\setlength{\fboxsep}{1pt}%
\begin{subfigure}{.12\textwidth}
  \includegraphics[width=\columnwidth]{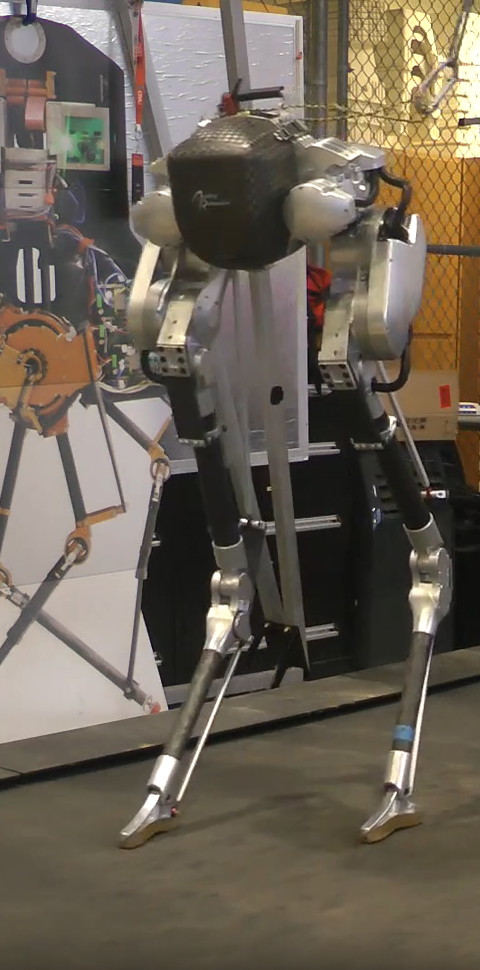}
\end{subfigure}%
\begin{subfigure}{.12\textwidth}
  \includegraphics[width=\columnwidth]{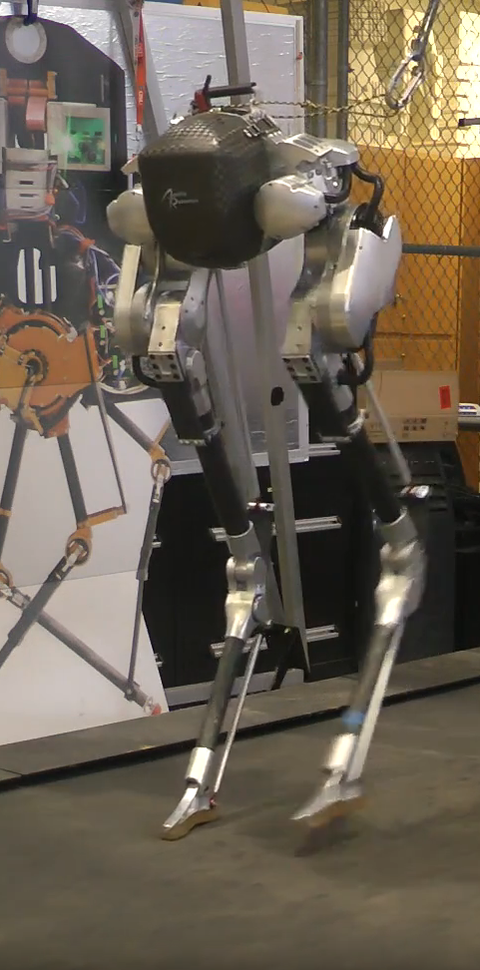}
\end{subfigure}%
\begin{subfigure}{.12\textwidth}
  \includegraphics[width=\columnwidth]{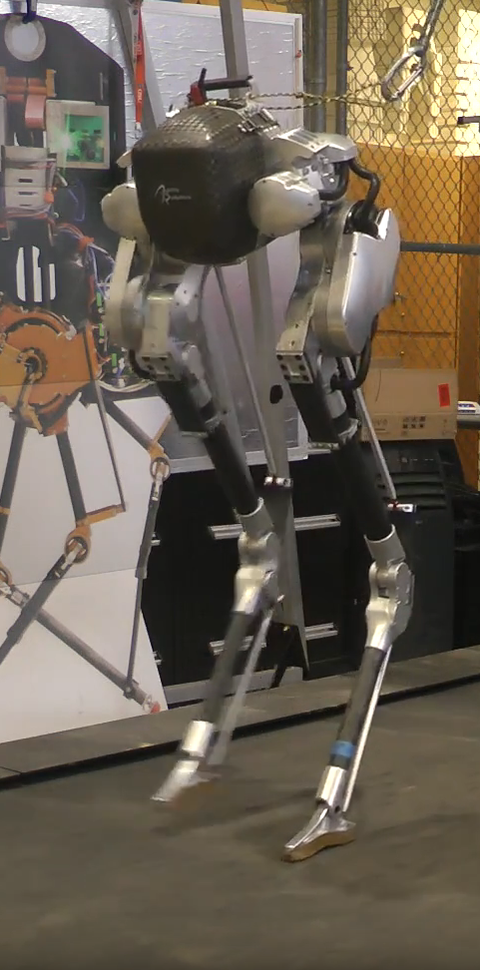}
\end{subfigure}%
\begin{subfigure}{.12\textwidth}
  \includegraphics[width=\columnwidth]{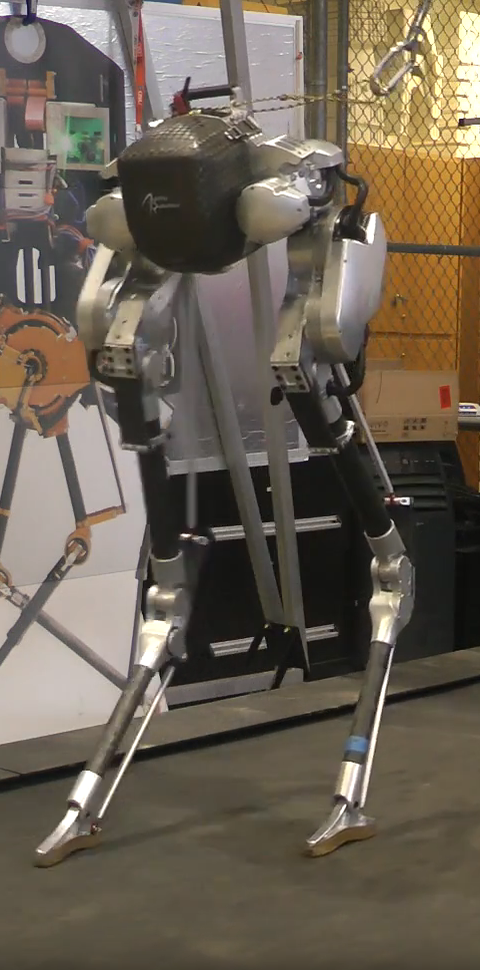}
\end{subfigure}%
\caption{Cassie walking on a treadmill with a neural network policy.}
\label{fig:cassie}
\end{figure}

In this paper, we propose a DRL design process that reflects and supports the
iterative nature of control policy design. At its heart is a data collection technique that 
allows us to recover a trained policy from a relatively small number of samples. 
With this technique, we can quickly compress and combine locomotion policies  
with supervised learning. By using policy gradient updates
that combine the supervised learning samples and conventional DRL policy-gradient samples, 
we allow for the iterative design of improved policies using new reward functions that encourage desired behaviors. 
We validate our approach in simulation and on a physical Cassie robot,
demonstrating stable walking policies with different styles at various speeds. 
Frames from a learned forward-walking gait for Cassie are shown in Fig.~\ref{fig:cassie}.

To summarize, this paper makes the following contributions:
\begin{itemize}

\item We present a simple-yet-effective technique to reconstruct policies from only a small number of samples, 
and show that robust variable-speed walking policies can be achieved on physical hardware using 
datasets of 5-10k tuples taken from simulations.

\item We combine reinforcement learning with supervised learning from this small number of samples. 
Guided by these samples, new policies can be learned by designing new reward functions 
that define desired changes to the behaviors while staying close to the original policy. 
This offers a strong alternative to "fine-tuning" approaches, where an existing policy
may be adapted via small changes and additions to an existing reward function, 
but which results in ever-more cumbersome reward functions and may exhibit unexpected changes in behavior.

\item We apply this approach to train various locomotion policies on a simulated model of the bipedal robot Cassie. These policies are successfully transferred to the physical robot without any further tuning or adaptation.

\end{itemize}

To the best of our knowledge, we believe this is the first time that neural network policies for 
variable-speed locomotion have been successfully deployed on a human-scale $3D$ bipedal robot such Cassie. 
The policies trained in simulation are directly transferred to the physical robot
without the use of the dynamics randomization methods. The gaits are comparable or faster in speed
than other gaits reported in the literature for Cassie, e.g.,~\cite{2018-arxiv-cassie}.

%% file: related.tex
\section{Related Work}

\subsection{Supervised Learning for Trajectory Optimization}
There exists significant prior art on generating optimal trajectories using supervised learning. 
This can be made fast and robust by precomputing a library of solutions and using a ``warm start'' 
for new problems using nearest neighbors, e.g., \citet{2013-robotica-trajectory_library,2017-iros-data_driven_control}.
Regression using neural networks has been used to generate optimal trajectories for bipedal 
robots~\cite{2017-icra-supervised_atrias} and quadrotors~\cite{2018-iros-learning_trajectories}. 
Guided Policy Search~\cite{2013-icml-gps} makes use of solutions from trajectory optimization to guide the policy search. 

\subsection{Imitation Learning}
Imitation learning seeks to approximate an expert policy. 
In its simplest form, one can collect a sufficiently large sample of state-action pairs from the expert and 
apply supervised learning, which is also referred to as behavior cloning, as used in early seminal
autonomous driving work by ~\citet{1988-nips-ALVINNAA}.
However, due to issues of compounding errors and covariate shift, 
this method often leads to failure \cite{2010-ICAIS-covariate_shift}. 
The DAGGER method~\cite{2011-AISTATS-dagger} is proposed to solve this problem, 
where the expert policy is iteratively queried to augment the expert dataset. 
\citet{2017-corl-dart} injected adaptive noise into the expert policy to reduce expert queries.
Recent work learns a linear approximation of the expert policy~\cite{2018-arxiv-neural_primitive}. 
Another line of work is to formulate imitation learning problems as reinforcement learning problems 
by inferring the reward signal from expert demonstration using methods such as GAIL~\cite{2016-arxiv-gail}. 
Expert trajectories can also be stored in a reinforcement learning agent's experience buffer to 
accelerate the reinforcement learning process~\cite{2018-arxiv-Q_learning_demonstration,2018-iros-acquire_nav}. 
Dynamic Movement Primitives~\cite{2003-iros-DMP,2004-RAS-lfd_biped} provide another approach 
to incorporating expert demonstration to learn motor skills.

\subsection{Distillation}
Supervised learning is often used to combine multiple policies. For example, 
it is successfully used to train policies to play multiple Atari games~\cite{2015-arxiv-distill,2015-arxiv-actor_mimic}.  
More recently, \citet{2018-ICLR-distill} use it to train a simulated 2D humanoid to traverse different 
types of terrains. These methods still suffer from the covariate shift problem and need to use DAGGER in the process.

\subsection{Bipedal Locomotion}
Bipedal locomotion skills are important for robots to be able to traverse terrains that are typical in human environments. 
Many methods use the Zero Moment Point (ZMP) to plan stable walking motions, e.g.,~\cite{1998-icra-asimo,2015-humanoids-ZMP_atlas}. 
Low dimensional models such as linear inverted pendulum (LIP) and spring loaded inverted pendulum (SLIP) 
can be used to simplify the robot dynamics~\cite{2001-iros-LIP,2018-rss-fast_online_traj_cassie,2018-humanoids-coupling} 
for easier and faster planning. To utilize the full dynamics of the robots, offline trajectory optimization 
such as direct collocation~\cite{2018-arxiv-cfrost} is often used to generate trajectories, 
and tracking controllers based on quadratic programming~\cite{2016-icra-constrained_systems} or 
feedback linearization~\cite{2018-arxiv-cassie} can be designed along these trajectories. 

Reinforcement learning has also been applied to bipedal locomotion, results on hardware are demonstrated 
on either 2D bipeds \cite{2010-iros-2dleo,2018-arxiv-rl_on_atrias} or bipeds 
with large feet~\cite{2004-iros-policy_gradient_biped}. 
More recently, deep reinforcement learning has been applied to 3D bipedal locomotion problems~\cite{2017-TOG-deepLoco,2018-SIGGRAPH-symmetry,2017-arxiv-parkour}. 
However, these works have not yet shown results on a physical robot.

%% file: preliminaries.tex
\section{Preliminaries}
In this section we briefly outline the reinforcement learning and imitation learning framework.
\subsection{Reinforcement Learning}
In reinforcement learning, we wish to learn an optimal policy for a Markov Decision Process (MDP). 
The MDP is defined by a tuple $\{\mathcal{S}, \mathcal{A}, P, r, \gamma\}$, where $S \in \mathbb R^n, A \in \mathbb R^m$ are the state space and action space of the problem, the transition function $P: S \times S \times A \to [0, \infty)$ is a probability density function, with $P(s_{t+1} \mid s_t, a_t)$ being the probability density of $s_{t+1}$ given that at state $s_t$, the system takes the action $a_t$. The reward function $r: S\times A \to \mathbb R$ gives a scalar reward for each transition of the system. $\gamma \in [0, 1]$ is the discount factor. The goal of reinforcement learning is to find a policy $\pi$, parameterized by $\theta$, where $\pi_\theta: S \times A \to [0, \infty)$ is the probability density of $a_t$ given $s_t$ that solves the following optimization problem:

\begin{align*}
\mathop{\mathrm{max}}_\theta J_{rl}(\theta) = &E\left[\sum_{t=0}^{\infty}\gamma^t{r({s}_t, {a}_t)} \right] \\
\text{subject to    } & s_{t+1} \sim P(. \mid s_t, a_t) \\
& a_t \sim \pi_\theta(. \mid s_t)
\end{align*}

Policy gradient algorithms~\cite{1999-nips-policy_gradient} are a popular approach to solving this problem, where $\nabla_\theta J_{rl}$ is estimated using on-policy samples, i.e., using data collected from the current stochastic policy.

\subsection{Imitation Learning}
In imitation learning, we have an MDP as defined above, and an expert policy $\pi_e$ is given. The goal of imitation learning is to find a parametrized policy $\pi_\theta$ that minimizes the difference between $\pi_\theta$ and $\pi_e$. More formally, we aim to solve the following optimization problem:

\begin{align}
\mathop{\mathrm{min}}_\theta J_{imit}(\theta) = &E_{s\sim p_e(s)}[(a - a_e)^2] \label{eq:imitation_obj} \\
\text{subject to    } & a \sim \pi_\theta(. \mid s) \nonumber \\
 & a_e \sim \pi_e(. \mid s) \nonumber
\end{align}

where $p_e(s)$ is the probability density of $s$ with policy $\pi_e$:
$$p_e(s) = \sum_{i=0}^\infty \gamma^i p(s_t=s \mid \pi_e)$$

The expectation in the objective is often estimated by collecting a dataset of expert demonstrations. In behavior cloning, the expert policy is assumed to be deterministic. This causes the well-known covariate shift problem, where the student policy will accumulate errors overtime and eventually drift to states that were not seen by the expert during data collection. Popular remedies to this issue include DAgger~\cite{2011-AISTATS-dagger} and DART~\cite{2017-corl-dart}, which query the expert policy iteratively to augment the dataset.

%% file: methods.tex
\section{Methods}
In this section, we present our method for collecting state-action pairs as a dataset for imitation learning, and how this dataset can be used to combine imitation learning and reinforcement learning.
In our iterative-design framework, we will consider a previously-learned policy as being the expert
for the next iteration of policy optimization.

\subsection{Data Collection}
If we assume $\pi_e(.\mid s)$ and $\pi_\theta(.\mid s)$ are Gaussian distributions with the same covariance, minimizing the imitation objective function~(\ref{eq:imitation_obj}) is equivalent to minimizing $J(\theta) = E_{s \sim p_e(s)}[(m_e(s) - m_\theta(s))^2]$, where $m_e, m_\theta$ are the means of $\pi_e$ and $\pi_\theta$. It is generally impractical to calculate this expectation exactly; in practice, we will collect an expert dataset $\mathcal{D} = \{(s_i,m_e(s_i))\}_{i=1}^N$ of size $N$, where $s_i$ is the state visited by the expert during policy execution, and minimize training error over $\mathcal{D}$, i.e, we will solve the following supervised learning problem: 
\begin{align}
\mathop{\mathrm{min}}_\theta J_{sp}(\theta) = &E_{s\sim \mathcal{D}}[(m_\theta(s) - m_e(s))^2]\label{eq:imitation_training_obj}
\end{align}
Note that during data collection, while we are recording only the mean of the policy, we are simulating a stochastic policy by adding noise to the mean during execution. This is related to \cite{2017-corl-dart}, where adaptive noise is added to the expert policy to prevent covariate shift. In our setting, since we already know the distribution of our expert policy, we can avoid iteratively querying the expert policy with adaptive noise, and just query the expert once at the beginning. 
We refer to this data collection method as Deterministic Action Stochastic State (DASS), 
since we only collect a deterministic actions, but at states that are sampled from the stochastic policy. 
Algorithm~\ref{alg:dass} summarizes our data collection procedure.

\begin{algorithm}
\caption{DASS}\label{alg:dass}
\begin{algorithmic}[1]
\State Initialize $\mathcal{D} = \{ \}$
\State Reset from some initial state distribution $s_0 \sim p_0(.)$
\For{$i=0,1\ldots, N$}
\State $\mathcal{D} = \mathcal{D} \cup \{(s_i, m_e(s_i)\}$
\State $a_i \sim \pi_e(. \mid s_i)$, $s_{i+1} \sim P(. \mid s_i, a_i)$
\If{$s_{i+1} \in \mathcal{T}$ for some termination set $\mathcal{T}$}
\State $s_{i+1} \sim p_0(.)$
\EndIf
\EndFor
\end{algorithmic}
\end{algorithm}

\begin{figure}[tb]
\includegraphics[width=4cm]{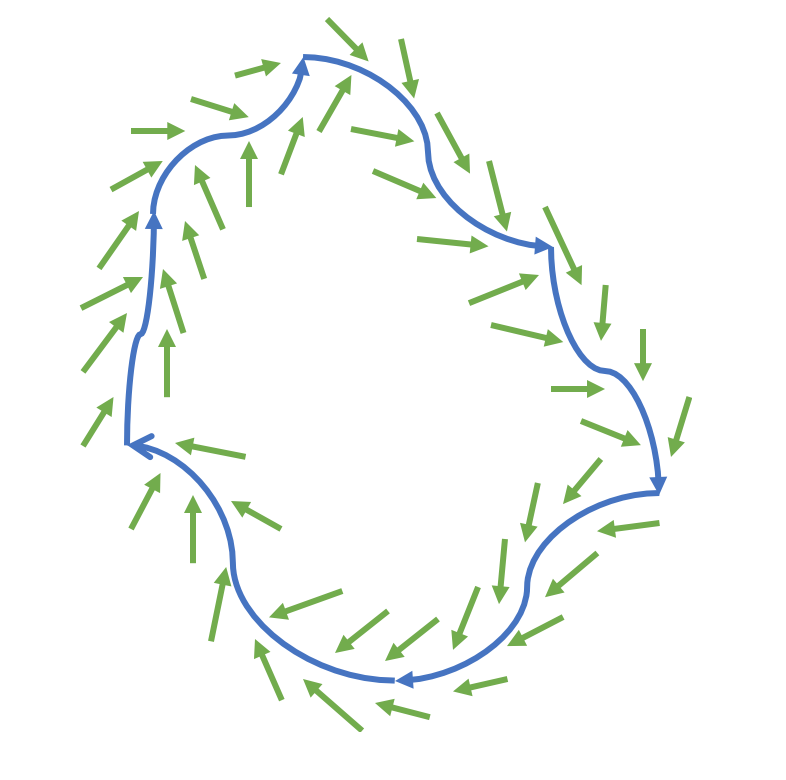}
\centering
\caption{A walking policy produces a limit cycle, represented by the blue closed curve, and the green arrows indicate the required feedback to return to the limit cycle.
}
\label{fig:limit_cycle}
\end{figure}

From a control perspective, this method for collecting expert data can be interpreted as follows. 
For policies such as walking that produce a limit cycle trajectory, 
recording the actions of an expert with no noise, i.e., just using the deterministic mean actions, 
then the collected data only covers the limit cycle and the student will not observe the feedback 
that should be applied when the state is outside of the limit cycle. With the noise of the stochastic policy, 
the expert is further able to provide data on how to return to the limit cycle from states not on the cycle, 
and the student will be able to learn a feedback controller along this limit cycle. 
This is illustrated schematically in Fig.~\ref{fig:limit_cycle}, where the blue curves represent 
the limit cycle produced by a deterministic policy, and the green arrows 
represent the deterministic feedback actions associated with the additional states 
resulting from the execution of the stochastic policy.

A key advantage of representing policies using the DASS tuples is that they can be combined in order to 
distill multiple specialized policies into one single policy.  
If we assume the desired skill specification is implicit in the state information, we can collect datasets $\mathcal{D}_i$ corresponding to multiple experts $\pi_{e_i}$, and use the union of these datasets $\mathcal{D} =  \cup_i \mathcal{D}_i$ for supervised learning.

\subsection{Combining Reinforcement Learning and Imitation Learning}
Imitation learning with DASS provides us a sample-efficient method to recover and combine expert policies. 
However, a realistic design process necessitates further iteration, where we wish to train 
policies that are further refined with respect to some criteria, while also remaining 
close to the original expert policies. To achieve this goal, we will add a constraint in the original formulation of the reinforcement learning problem:
\begin{align*}
\mathop{\mathrm{max}}_\theta J_{rl}(\theta) = &E\left[\sum_{t=0}^{\infty}\gamma^t{r({s}_t, {a}_t)} \right] \\
\text{subject to    } & s_{t+1} \sim P(. \mid s_t, a_t) \\
& a_t \sim \pi_\theta(. \mid s_t) \\
& J_{sp}(\theta) \leq \epsilon
\end{align*}
To make this problem easier, we make the constraint on $J_{sp}$ a soft constraint and rewrite 
the objective to be $J_{total} = J_{rl} - wJ_{sp}$. At each iteration, we will 
estimate $\nabla_{\theta_t} J_{rl}$ using the usual policy gradient algorithm, and update $\theta$ 
according to $\theta_{t+1} = \theta_t + \alpha(\nabla_{\theta_t} J_{rl} - w\nabla_{\theta_t} J_{sp})$.

Note that the reward function $r$ need not be related to the expert skill. 
If we set $r(s,a) = 0,  \forall s\in S, \forall a\in A$, then we recover the imitation learning problem. 
Furthermore, we can learn new skills while not forgetting expert skills. 
For example, the expert can be a policy for a robot walking forward while $r$ is rewarding the robot to walk backward. 
If one has reason to believe that the expert policy is suboptimal, we can also use this method to 
fine-tune the expert policy by defining $r$ to be the same as what was used to train the expert policy. 
The benefit of this is that we don't need access to the expert policy for the fine-tuning to happen. 
Finally, we can design rewards so that the new policy satisfies additional specific objectives that we desire, 
such as smoother movement or lifting the feet higher at each step.

%% file: set_up.tex
\section{Experimental Setup}
\subsection{Robot Specification}
\begin{figure}[tb]
\centering
\setlength{\fboxsep}{1pt}%
\begin{subfigure}{.20\textwidth}
  \includegraphics[width=\columnwidth]{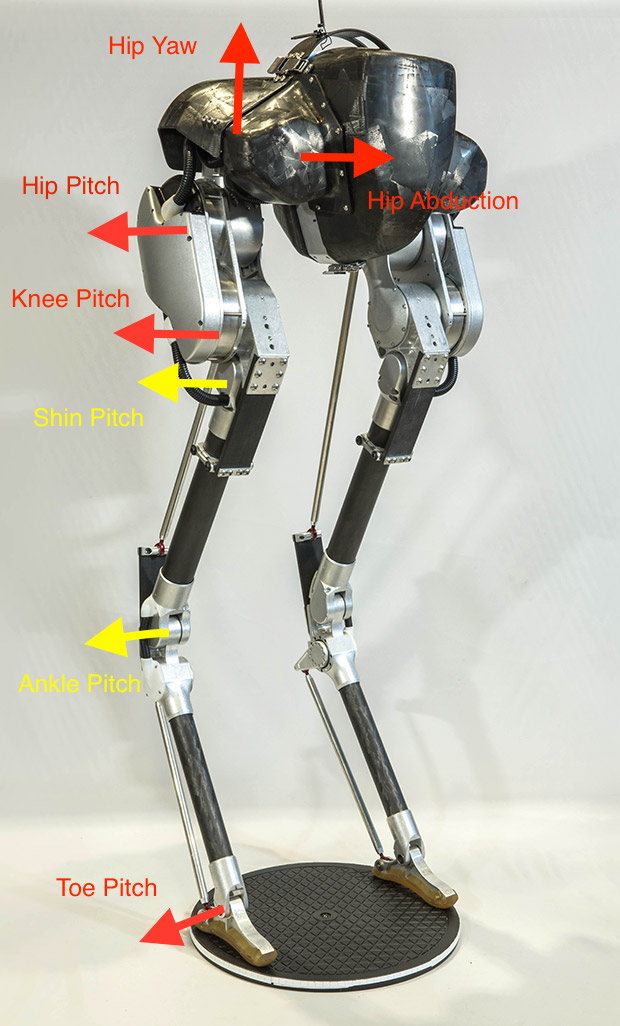}
\end{subfigure}%
\begin{subfigure}{.25\textwidth}
  \includegraphics[width=\columnwidth]{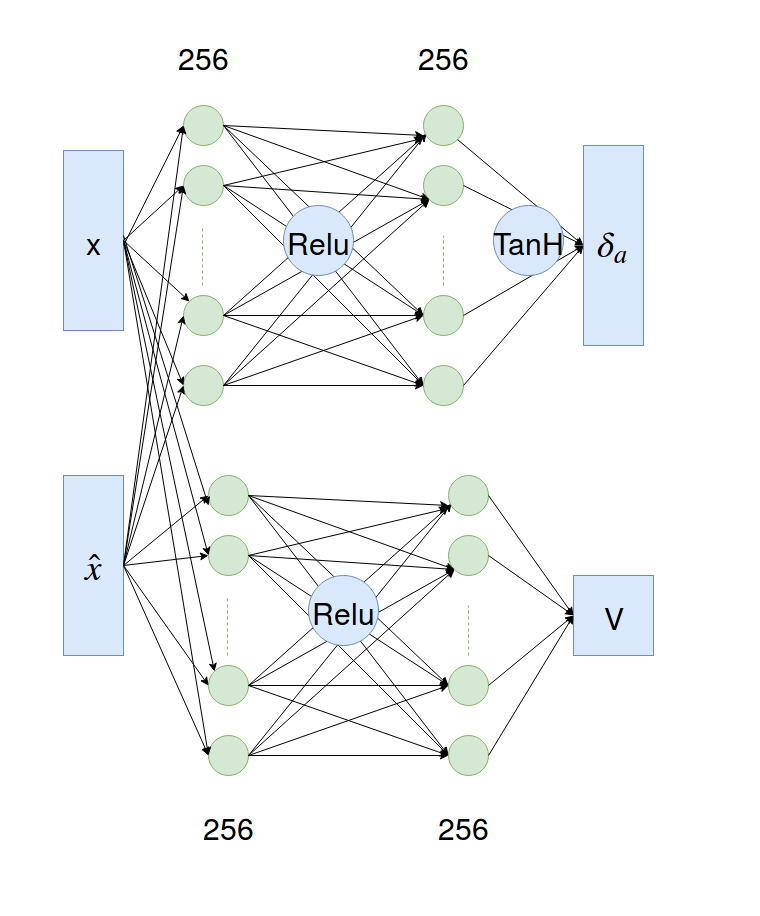}
\end{subfigure}%
\caption{Left: The bipedal robot Cassie used for evaluation. The red arrows indicate the axes of actuated joints, the yellow arrows indicate passive joints with stiff leaf springs attached for compliance. Right: The neural network used to parameterize the policy.
}
\label{fig:framework}
\end{figure}

We evaluate our methods using the Cassie bipedal robot. 
Cassie, shown in Fig.~\ref{fig:framework}, is designed and built by Agility Robotics. 
It stands approximately 1 meter tall and has a total mass of 31 kg, 
with most of the weight concentrated in the pelvis. 
There are two leaf springs on each leg to make them more compliant. 
This introduces extra underactuation into the system, and it makes the control design more difficult 
for traditional techniques.

\begin{figure*}[tbh!]
\centering
\setlength{\fboxsep}{1pt}%
\begin{subfigure}{.80\textwidth}
  \includegraphics[width=\columnwidth]{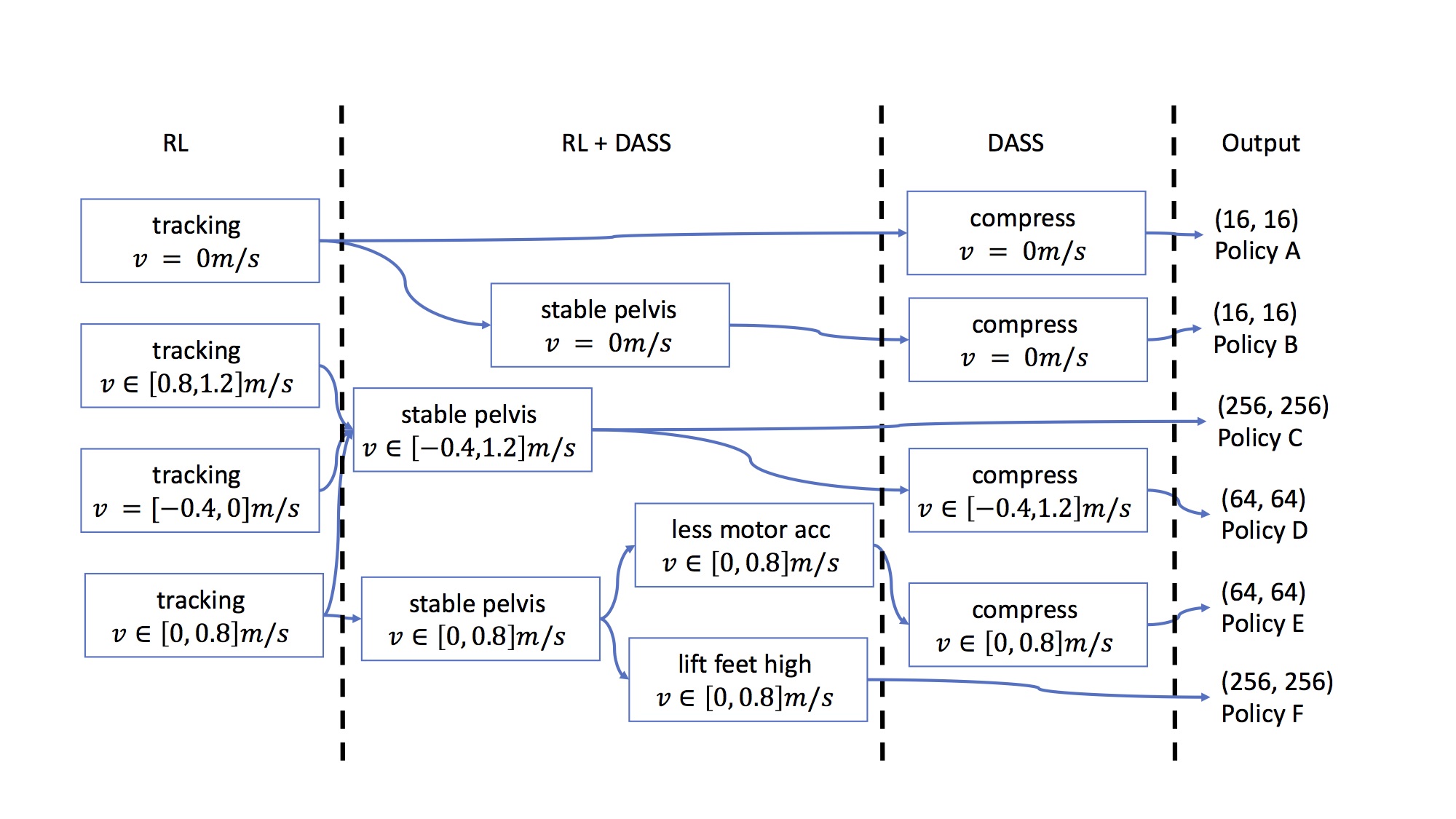}
\end{subfigure}%
\caption{Our policy design process. Four tracking-based policies are used as a starting point. 
DASS samples are passed from one policy to the next according to the arrows. 
}
\label{fig:policy_list}
\end{figure*}

\subsection{Training Framework}
We adopt the framework used in \cite{2018-IROS-cassie} for training several initial policies $\pi_e$, 
where we reward the agent for producing motion that approximately reproduces a set of specified reference motions. 
The input state to the  policy is given by $S = \{X, \hat{X}\}$, where $X$ is the state of the robot that evolves according to the robot's dynamics, and $\hat{X}$ is the reference motion of the robot that evolves deterministically according to the motion we desire to track. The state of the robot includes the height, orientation expressed as a unit quaternion, velocities, angular velocities and acceleration of the pelvis, joint angles and joint velocities of each joint. In total, this gives us a $85D$ input vector. We use the commonly-adopted Gaussian Policy as output, where the neural network will output the mean of the policy and Gaussian noise is injected on top of the action during execution. 
The output and the reference motion are summed to produce target joints angles for a low level PD controller.
Instead of making the covariance of the Gaussian policy a learnable parameter, we use a fixed covariance for our policy. 
We assume that the Gaussian distribution in each dimension is independent, with a standard deviation of $exp(-2) \approx 0.1$ radians. A benefit of the fixed covariance is that because of the noise constantly injected into the system during training, the resulting policy will adapt itself to handle unmodeled disturbances during testing, as demonstrated in previous work~\cite{2018-IROS-cassie, 2018-TOG-deepMimic}. The network architecture is shown in Fig.~\ref{fig:framework}. 
The policy is trained with an actor-critic algorithm using a simulated model of Cassie with 
the MuJoCo simulator~\cite{2012-iros-mujoco}, with the gradient of the policy estimated using Proximal Policy Optimization \cite{2017-arxiv-ppo}. The simulator includes a detailed model of the robot's rigid-body-dynamics, 
including the reflected inertia of the robot's motors, as well as empirically measured noise and delay 
for the robot's sensors and actuators.

\subsection{Policy Training}

The design process we use for training our policies is summarized in Fig.~\ref{fig:policy_list}.
We initially train four different tracking-based policies: stepping in place; 
walking forward with speed ranging from $v \in [0, 0.8]m/s$; 
walking backward with $v \in [-0.4, 0]m/s$; and 
fast walking forward with $v \in [0.8, 1.2]m/s$. 
The reference motions for the stepping in place and walking forward at $0.8m/s$ motions are recorded 
from motion produced by an existing heuristically tuned controller, 
and the reference motions for walking at other speeds are obtained by scaling the 
translation and velocity of the walking forward motion by the desired value. 
There exist numerous other  choices for obtaining reference motions, including using direct collocation \cite{2018-arxiv-cfrost} or key framing, but this is beyond the scope of this paper. 
The reference motions we work with are symmetric and the robot itself is nearly symmetric, 
and thus it is natural to enforce symmetry in the policies as well. 
We adopt a similar approach to~\cite{2018-TRO-gait_opt}, where we transform the input and output every 
half walking cycle to their symmetric form. During training, we apply reference state initialization 
and early termination techniques as suggested by \citet{2018-TOG-deepMimic}, where each rollout is started from some states sampled from the reference motions and is terminated when the height of the pelvis is less than $0.4$ meters, 
or whenever the reward for any given timestep falls below a threshold of $0.3$.

The initial tracking based policies are then used as the starting point for further design exploration. 
We show $6$ of these policies running on the physical robot in our supplementary video. 
Several intermediate policies are also successfully tested on the robot,
but are not shown due to video-duration constraints.
At each level, all policies are trained from scratch instead of fine-tuning the previous policies. 
This is important for distilling multiple policies together, and for policy compression
on to a smaller network; in these cases, an original policy will not be available.
We further note that fine-tuning a policy based on a new reward function often results
in undesired changes to the policy as it can readily "forget" the objectives and motion features
of the starting policy.

\subsection{Hardware Tests}
We deploy a selection of trained policies on a physical Cassie robot. 
The state of the robot is estimated using sensor measurements from an IMU and joint encoders, 
which are fed into an Extended Kalman Filter to estimate parts of the robot's state, 
such as the pelvis velocity. This process runs at 2 kHz on an embedded computer directly connected to the robot's hardware.
This information is sent over a point-to-point Ethernet link to a secondary computer onboard the robot, 
which runs a standard Ubuntu operating system and executes the learned policy using the PyTorch framework. 
The policy updates its output joint PD targets once every 30 ms based on the latest state data and sends the targets 
back to the embedded computer over the Ethernet link. The embedded computer executes a PD control loop for each 
joint at the full 2 kHz rate, with targets updating every 30 ms based on new data from the policy execution.

Rapid deployment and testing is aided by the simulator using the same network-based interface as the physical robot, 
which means that tests can be moved from simulation to hardware by copying files to the robot's onboard computer 
and connecting to a different address. The robot has a short homing procedure performed after powering on, 
and can be left on in between testing different policies. The same filtering and estimation code as used on hardware 
is used internally in the simulator, rather than giving the policy direct access to the true simulation state. 
The network link between two computers introduces an additional 1-2 ms of latency beyond running the simulator 
and policy on the same machine, and many of the robot's body masses are slightly different from the 
simulated robot due to imprecisely modeled cabling and electronics and minor modifications made to the robot 
since the simulation parameters were produced.

%% file: compress_distill.tex
\section{Policy Compression and Distillation}
In this section, we present results for using DASS to compress and distill multiple policies. 
In the experiment, we update the student policies using ADAM \cite{2014-arxiv-adam} with the 
supervised loss from Equation \ref{eq:imitation_training_obj} with a batch size of $128$. 
We collect an additional $300$ DASS samples for evaluating validation error. 
We stop the training when the training error improves less than $10^{-5}$ over $1000$ iterations.

\subsection{Policy Compression}

\begin{figure}[tb]
\includegraphics[width=8cm]{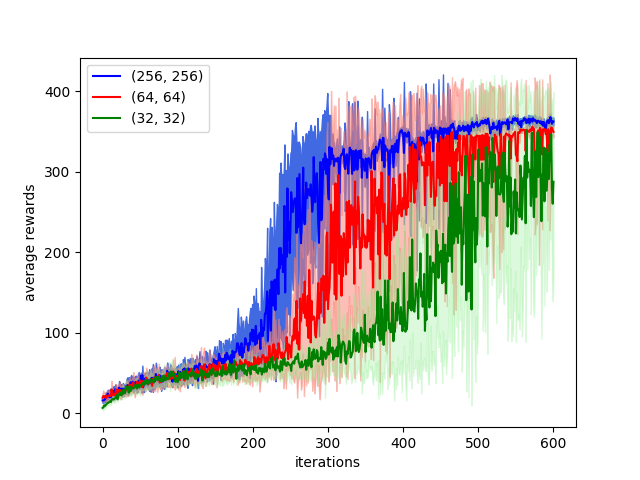}
\caption{Network sizes impact the final result for reinforcement learning. We observe that larger network sizes typically learn faster and yield more stable policies. Compared to the $(256, 256)$ network, 
the learning proceeds much more slowly for network sizes of $(64, 64)$ and $(32, 32)$, 
and has a larger variance, indicating the final policy is not robust to noise.
}
\label{fig:size_comparison}
\end{figure}

In deep reinforcement learning, network size often plays an important role in 
determining the end result~\cite{2018-AAAI-DRL_matters}. It is further shown in~\cite{2015-arxiv-distill}  
that for learning to play Atari game, a large network is necessary during RL training to achieve good performance, 
but that the final policy can be compressed using supervised learning without degrading the performance. 
For our problem, we also observe that using a larger network size for reinforcement learning improves 
learning efficiency as well as producing more robust policies, as shown in Fig.~\ref{fig:size_comparison}.

\begin{figure}[t]
\includegraphics[width=8cm]{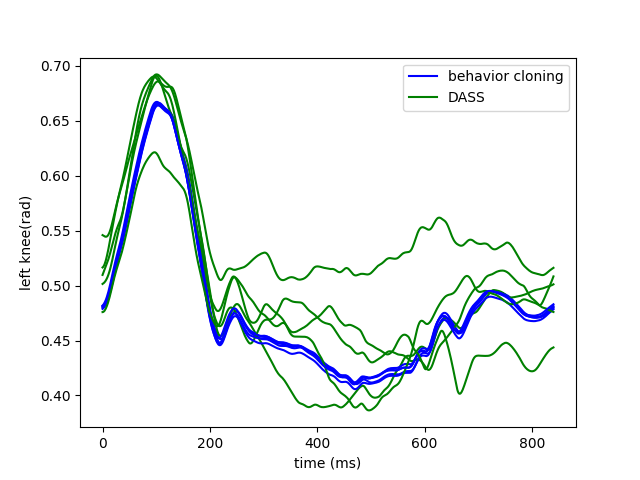}
\caption{Joint angles of the left knee in the expert (teacher) dataset, as collected via policy cloning or DASS. 
Behavior cloning only visits a limited set of states, namely those very near the limit cycle.
}
\label{fig:deterministic_vs_stochastic}
\end{figure}

While we need a large network to efficiently do reinforcement learning, we find that 
we can compress the expert policy into a much smaller size network while maintaining the robustness 
of the final policy. With as little as $600$ samples, we can recover a stepping in place policy with 
a $(16, 16)$ hidden layer size. Fig.~\ref{fig:deterministic_vs_stochastic} compares a dataset collected using behavior cloning with that of the DASS collection strategy. 
Table~\ref{tab:policies_comparison} compares policies trained using supervised learning across 
varying choices of hidden layer sizes, numbers of training samples, and the presence or absence of noise
during data collection. With only $600$ samples, a large network can easily overfit the training data. 
We find that while larger networks can indeed have this issue, having validation error orders of magnitude 
larger than the training error, the resulting policy still performs comparably to the original policy
in terms of robustness.

\begin{table*}[tb]
  \begin{center}
    \begin{tabular}{l|c|c|c|c|c|c}
      \textbf{policy} & \textbf{training loss}&\textbf{validation loss}& \textbf{no noise}  & \textbf{0.1 policy noise} &\textbf{20\% mass noise} &\textbf{50N pushes} \\
      \hline
      expert & $0$ &$0$ & $394.53$ & $389.82$& $378.32$ &$367.28$\\
      300 samples &$1.20\pm 0.08 \times10^{-3}$&$2.99\pm 0.25 \times10^{-3}$ & $392.88\pm 0.34$  &$366.53 \pm 26.79$ & $350.36 \pm 15.08$ & $346.27 \pm 5.68$\\
      600 samples & $1.33\pm 0.20 \times10^{-3}$ &$2.08\pm 0.12 \times10^{-3}$ & $394.24\pm 0.48$ & $388.66 \pm 0.37$ & $375.03 \pm 3.49$ & $363.56 \pm 3.11$ \\
      $(512, 512)$ & $6.00 \pm 1.18 \times 10^{-6}$ & $5.45 \pm 0.42 \times 10^{-4}$ & $394.36 \pm 0.34$ & $389.49 \pm 0.34$ & $371.65 \pm 4.04$ & $351.52 \pm 19.00$ \\
      $(8, 8)$ & $5.16 \pm 0.42 \times 10^{-3}$ & $6.17 \pm 0.65 \times 10^{-3}$ & $92.52 \pm 20.94$ & $72.44 \pm 10.63$ & $60.69 \pm 9.64$ & $81.47 \pm 12.94$ \\
      no noise & $5.04 \pm 1.73 \times 10^{-4}$ & $8.11 \pm 0.94 \times 10^{-3}$ & $66.19 \pm 8.96$  &$50.66 \pm 4.79$ & $65.82 \pm 13.47$ & $66.59 \pm 6.00$
    \end{tabular}
    \caption{Comparison of policies trained with various settings. The default hidden layer size is $(16, 16)$. 
    We evaluate the robustness of each policy by injecting noise of varying magnitude to the policy actions, 
    increasing the mass of the pelvis by $20 \%$, and applying pushes of $50N$ in the forward direction for 
    $0.2$ second every $3$ seconds, and report the cumulative rewards each policy obtained over $400$ control steps.}
  \label{tab:policies_comparison}
  \end{center}
\end{table*}

We successfully test the $(16, 16)$ policy on the physical robot. It exhibits similar behavior as in simulation, 
with the robot stepping in place while supporting its own weight. 
However, the pelvis exhibits an undesirable shaky movement, both in simulation and on the physical robot,
shown in the supplementary video.
This corresponds to Policy A in Fig.~\ref{fig:policy_list}. 

\subsection{Policies Distillation}
After training a network for a skill, we may want the policy to learn additional skills. 
In the context of the Cassie robot, we desire a control policy to not only step in place, 
but to also walk forwards and backwards on command. However, catastrophic forgetting 
can occur when trying to learn new skills. Distillation is one way to deal with forgetting. 
We can learn policies that master these skills separately and then distill these policies 
into a single policy with supervised learning.
We distill the three expert policies that are trained separately for walking forward, stepping in place and walking backward into one policy that masters all three tasks. With $600$ samples collected from each of these policies, 
we are able to combine these policies into one policy with a hidden layer size of $(64, 64)$.

%% file: rl_il.tex
\section{Iterative Design with Changing Rewards}

\subsection{Stable Pelvis Movement}
As noted in the previous section, the tracking based policy results in an undesirable shaking of the robot body.
While in simulation this does not affect the ability of the robot to complete its task, 
this places excessive wear on the physical robot hardware. 
We now apply the framework that combines policy-gradient RL updates and DASS-based policy cloning,
in support of iterative policy design.
The reward is dedicated to achieving stable movement of the robot body while maintaining desired speed. 
Specifically, we set the rewards to be $r = 0.5\exp{(-\norm{\omega_{pelvis}}^2)} + 0.5r_{rp}$, 
where $\omega_{pelvis}$ is the angular velocity of the pelvis, and $r_{rp}$ ensures a policy that
tracks the desired velocity.

\begin{figure}[h]
\includegraphics[width=8cm]{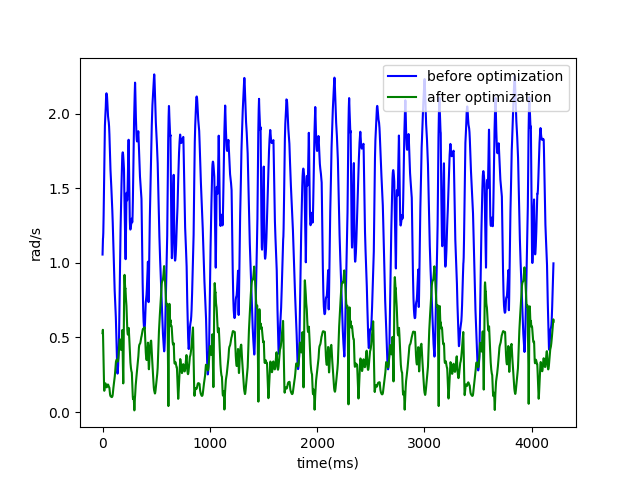}
\centering
\caption{Comparison of the norm of the angular velocity of the pelvis before and after optimization.
}
\label{fig:angular_comparison}
\end{figure}

We first experiment with the simple approach of simply fine-tuning the previous policy using 
the new reward, with a desired velocity of $0$. 
This results in the robot learning to stand still, and while this is a perfectly usable policy for this particular reward, 
it has effectively forgotten how to step in place. 
We next test learning that incorporates DASS samples into the policy update. 
To balance the number of DASS samples and on-policy samples,
on each iteration we train on 3000 DASS samples, using supervised learning and which are always drawn from the same set, 
and 3000 on-policy policy-gradient samples.
The resulting policy produces the desired stepping-in-place motion with much smoother pelvis movement than the 
original tracking based policy. Fig.~\ref{fig:angular_comparison} shows the comparison of the 
norm of the angular velocity of the pelvis before and after the optimization. 
We then compress this policy by collecting 600 new DASS samples from this policy and 
test it on the physical robot. This corresponds to Policy B in Fig.~\ref{fig:policy_list} and can be seen in the supplementary video. The physical robot is able to step in place with a markedly smoother motion. 

We extend this iterative-improvement approach to an tracking-based policy that is capable of walking at different speeds, 
from stepping in place to walking forward at a maximum speed around $0.8m/s$. 
This policy suffers from the same problem as the tracking-based stepping in place policy, 
where the pelvis is shaking at a moderately high frequency. 
For each speed, We similarly collect 3000 DASS samples and 3000 policy-gradient samples, for a total of 10 speeds,
$0m/s$ -- $0.8m/s$, taken at increments of $0.08m/s$.
The final policy produces stable walking motion that can be commanded at various speeds, 
both in simulation and on the physical robot.

We distill this policy and two other tracking-based policies that can walk backwards at $0.4m/s$ and 
forwards at $1.2m/s$ with the stable pelvis reward. As before, we collect samples for these policies 
at a increments of $0.08m/s$, each with $3000$ samples, and train a final unified policy that can 
walk at speeds from $-0.4m/s$ to $1.2m/s$ with stable pelvis movement. 
We test this on the physical robot, and the robot is able to achieve $1.14m/s$ on the treadmill
as well as slow backwards walks. This policy is then further compressed to a $(64, 64)$ hidden layer 
size network using supervised learning with DASS, with $600$ samples collected for each speed sampled at 
increments of $0.08m/s$. On the physical robot, this policy can achieve $0.8m/s$. 
We also compress this policy by collecting samples with speeds sampled at sparser increments of $0.4m/s$. 
The final policy has similar capabilities on the physical robot, although it is less responsive
to commanded changes in speed. The policies before and after compression correspond to Policy C and Policy D in Fig.~\ref{fig:policy_list} and in the video.

\subsection{Other Stylistic Reward}

\begin{figure*}[tbp]
\includegraphics[width=18cm]{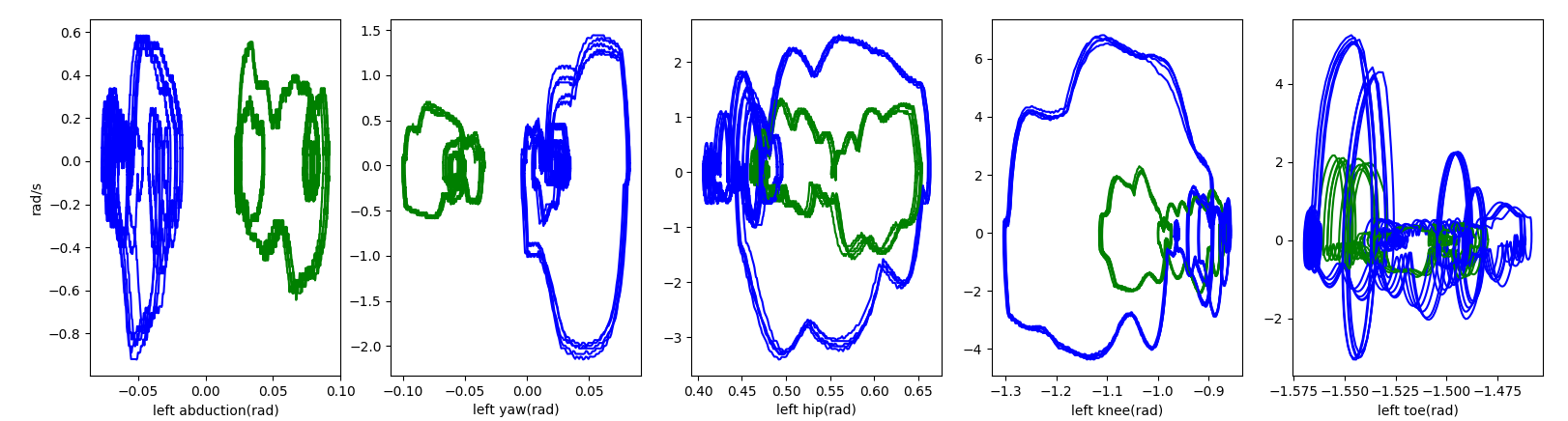}
\centering
\caption{Phase portrait for all the joints on the left leg during step in place. The blue curve is before optimizing for less joint accelerations, and the green curve is after.
}
\label{fig:phase_portrait}
\end{figure*}

We experiment with additional stylistic rewards.
We observe that the previous policies still exhibit noisy movement,
and we thus optimize (solely) for reduced joint accelerations. As before, transfer from the previous
policy is achieved using DASS sampling, which is then coupled with policy-gradient RL.
Fig.~\ref{fig:phase_portrait} compares the difference of the motion before and after the optimization. We then compress this policy to a policy with hidden layer of size $[64, 64]$ using the DASS samples.
We also experiment with a reward to lift the feet of the robot higher. 
The previous policies lifts the feet up to $10$~cm during each step. 
We penalize the policy for lifting the foot less than $20$~cm. 
Guided by DASS samples from the previous policy, the new policy 
learns to lift the feet up to $20$~cm while maintaining good walking motions.

\begin{figure*}[tbp]%
\centering
\setlength{\fboxsep}{1pt}%
\begin{subfigure}{.14\textwidth}
  \includegraphics[width=\columnwidth]{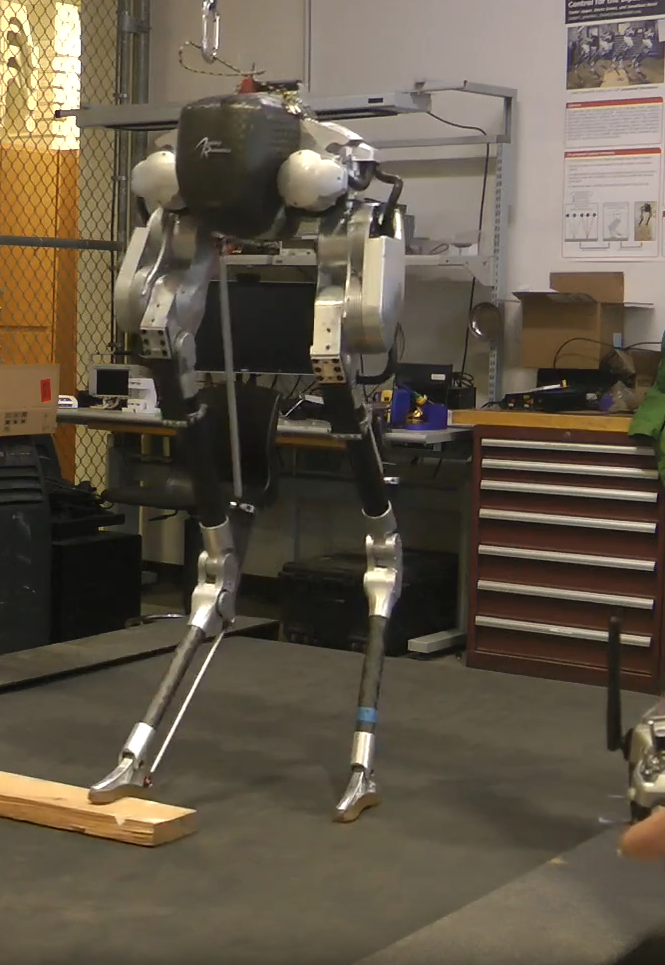}
\end{subfigure}%
\begin{subfigure}{.14\textwidth}
  \includegraphics[width=\columnwidth]{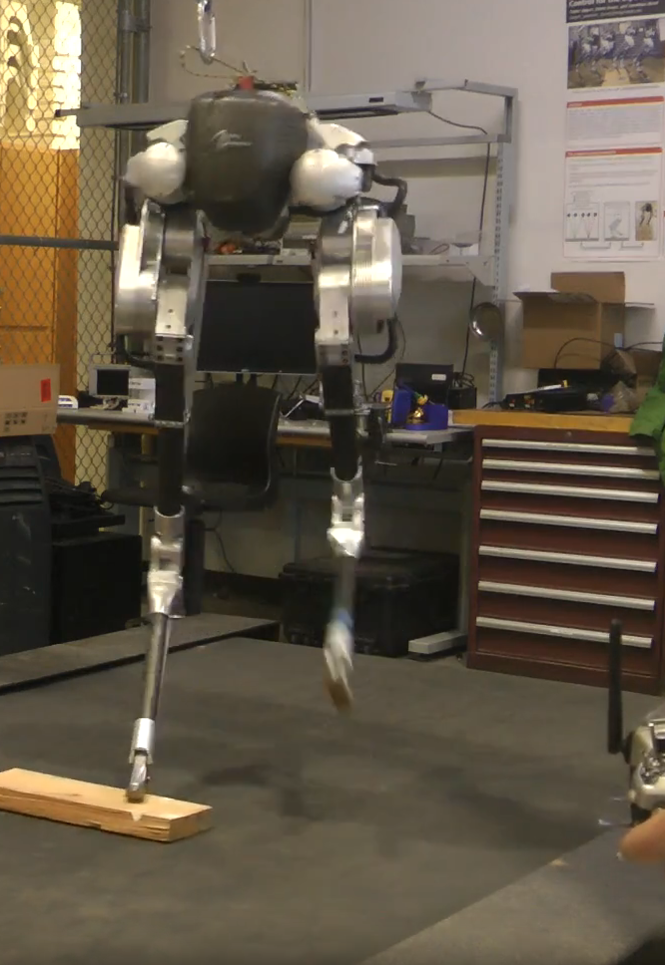}
\end{subfigure}%
\begin{subfigure}{.14\textwidth}
  \includegraphics[width=\columnwidth]{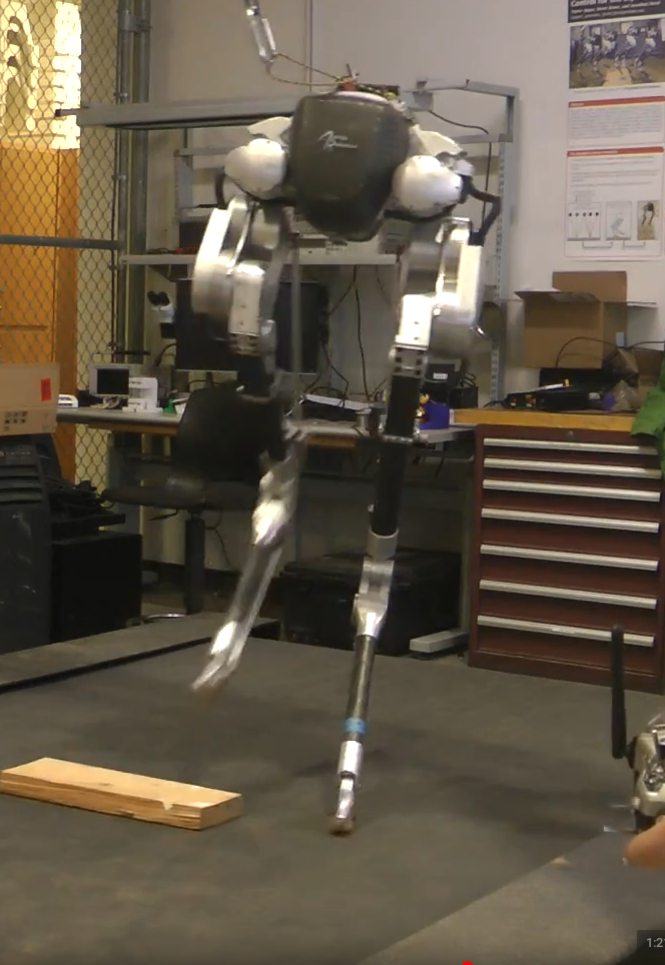}
\end{subfigure}%
\begin{subfigure}{.14\textwidth}
  \includegraphics[width=\columnwidth]{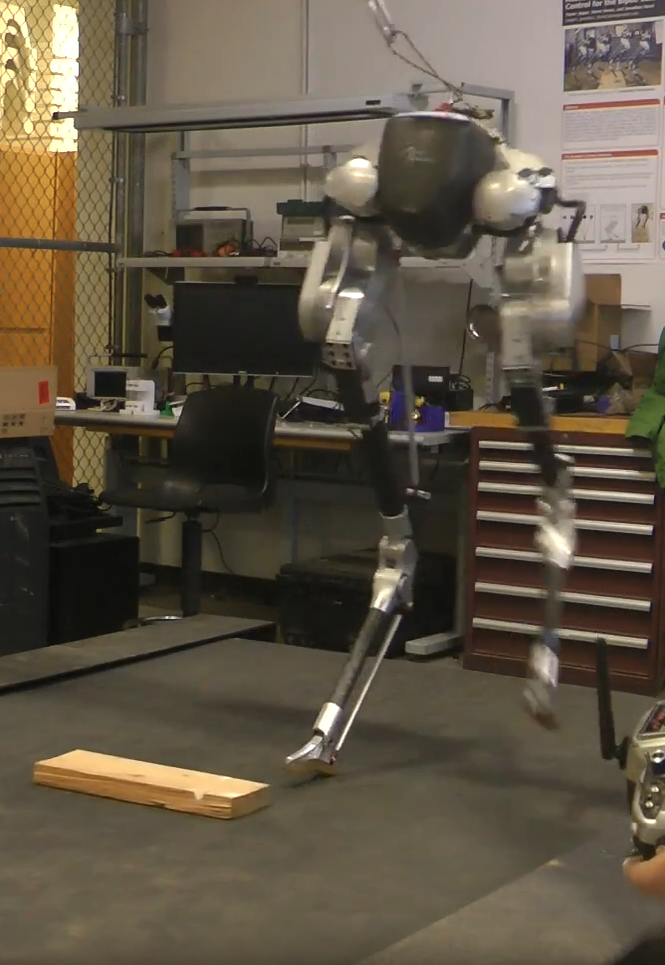}
\end{subfigure}%
\begin{subfigure}{.14\textwidth}
  \includegraphics[width=\columnwidth]{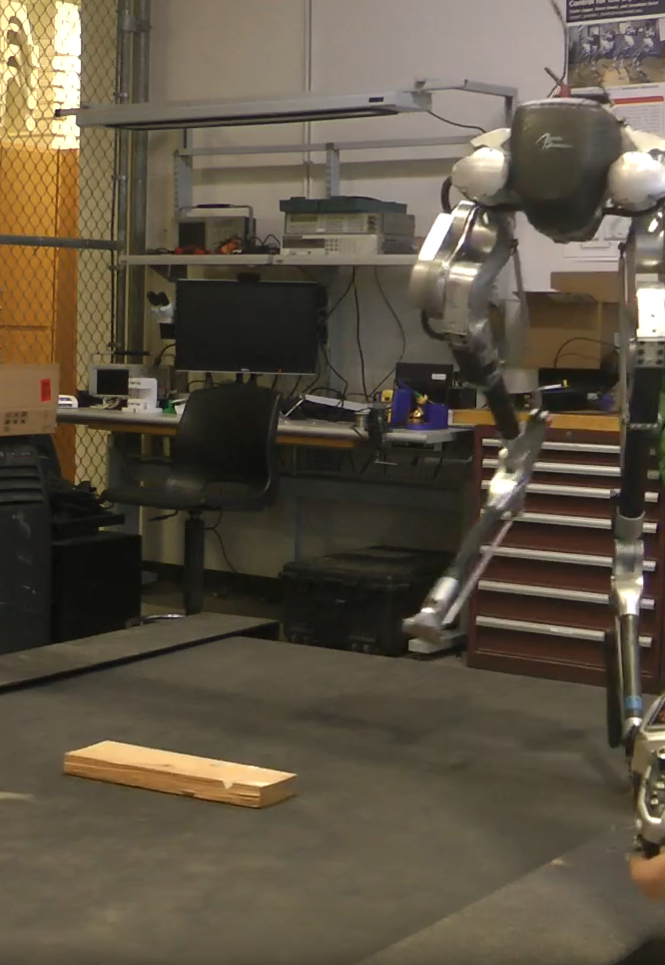}
\end{subfigure}%
\begin{subfigure}{.14\textwidth}
  \includegraphics[width=\columnwidth]{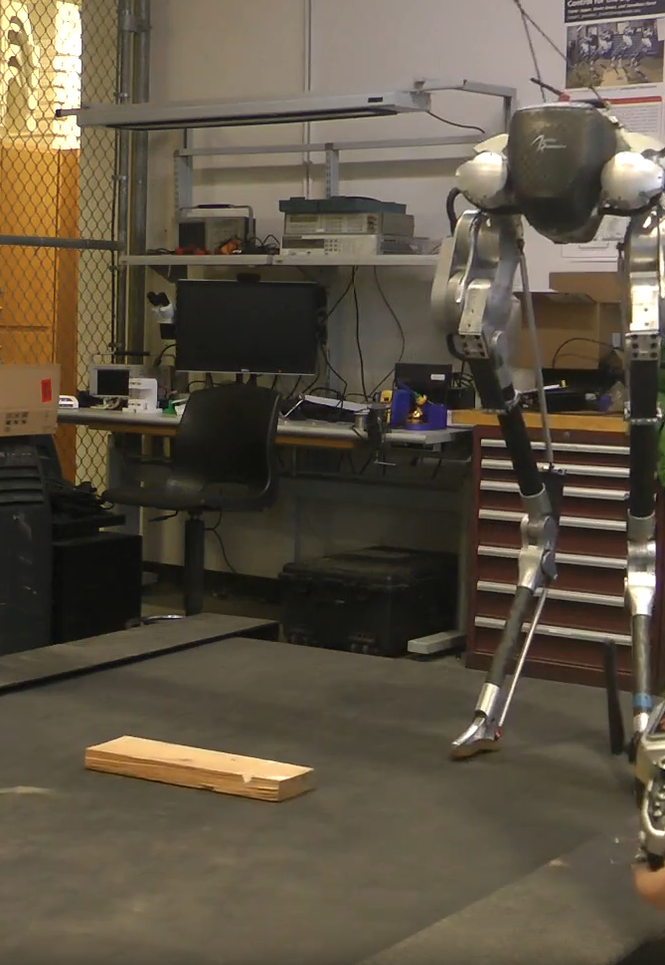}
\end{subfigure}%
\begin{subfigure}{.14\textwidth}
  \includegraphics[width=\columnwidth]{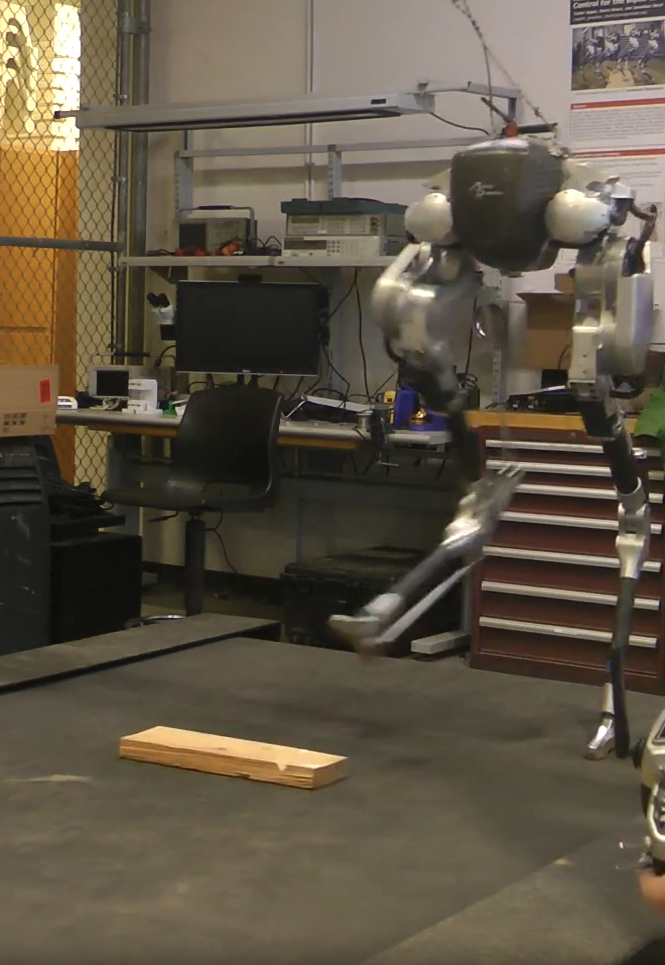}
\end{subfigure}%
\caption{Cassie recovers from stepping on an unexpected obstacle.}
\label{fig:cassie_vs_board}
\end{figure*}

We test these policies on the physical robot. 
The motions on the robot are comparable to the motions in simulation.
The policy that rewards low joint accelerations makes significantly softer and 
quieter ground contact. The policy optimized for lifting its feet higher achieves higher stepping. 
We further test the robustness of the high-stepping policy by placing boards 
in front of the robot. The policy is able to cope with this unmodeled disturbance and recover after several steps,
as shown in Fig.~\ref{fig:cassie_vs_board}. These policies correspond to Policy E and Policy F in Fig.~\ref{fig:policy_list} and are also shown in the video.

%% file: conclusions.tex
\section{Conclusion and Discussion} 

\label{sec:conclusion}
In this paper, we present a data collection technique (DASS) that enables us to quickly recover, 
compress and distill multiple policies using supervised learning. Importantly, we demonstrate
that DASS-based transfer learning can be integrated with policy-gradient RL methods.
This directly supports an {\em iterative design process}, where each iteration of the design
can optimize exclusively for a reward function that targets a desired change to the policy
or motion style. We validate this approach on the bipedal robot Cassie, 
achieving stable walking motions with different styles at various speeds.

The final policies obtained are robust to unmodeled noise and enable us to transfer them
from simulation to the physical robot without difficulty. This differs from most sim-to-real results, 
where a large range of dynamic randomization is often needed to ensure successful transfer, 
despite performing careful system identification~\cite{2018-rss-sim_to_real} and using quadrupedal systems 
that may have more passive stability than a human-scale biped. 

We show that the policies can be robust without resorting to dynamics randomization.
This is shown in simulation, where mass of the pelvis is perturbed by $20 \%$, 
and by the successful transfer to the physical robot, which exhibits changing dynamics during its operation cycle.
We hypothesize the robustness stems from learning stochastic policies that operate at a low control rate, 
allowing the final policies to adapt to other noise. 
It will be interesting to further identify what are the most important considerations that 
ensure sim-to-real success instead of always requiring dynamic randomization, 
which can cause the final policy to be overly-conservative.
We do note that the physical robot experiments exhibit asymmetric step lengths
to a degree that is not seen in simulation. The source of this remains to be determined,
but the policies are robust despite this difference. 

We show that while it is beneficial to use a relative large neural network 
during the reinforcement learning phase, the final policies can usually be represented by much smaller networks. 
It will be interesting to learn abstractions 
relevant to locomotion and to be able to reuse such abstractions for more efficient learning and planning.


An advantage of using deep neural networks is that they can be readily extended to develop policies
that directly accept rich perceptual input such as images, unlike traditional control methods.
We wish to give Cassie such visual input and use this in support of visual navigation. 

Our policies currently still take a reference motions as an input. However,  once the initial tracking policy
has been trained, the policy is free to develop its own movement styles according to the subsequent iterative
optimizations, and will even learn to stand still if given an appropriate reward function.
The reference motion does provide the policies a means to condition themselves on time or motion phase, 
similar to \cite{2018-TOG-deepMimic}. In some sense, this gives the policy more flexibility 
since with enough motions and offline training, it can be capable of generalizing to other unseen 
reference motions during testing.
In another sense, the reference motion poses constraints on the final motions, i.e., it may be more difficult 
for a motion to make timing adjustments. Given that pure state-based feedback can also yield
plausible locomotion, e.g., \cite{2018-SIGGRAPH-symmetry}, it will be interesting to seek a balance between these two representations.

%% file: paper_template.bbl
\begin{thebibliography}{43}
\providecommand{\natexlab}[1]{#1}
\providecommand{\url}[1]{\texttt{#1}}
\expandafter\ifx\csname urlstyle\endcsname\relax
  \providecommand{\doi}[1]{doi: #1}\else
  \providecommand{\doi}{doi: \begingroup \urlstyle{rm}\Url}\fi

\bibitem[Apgar et~al.(2018)Apgar, Clary, Green, Fern, and
  Hurst]{2018-rss-fast_online_traj_cassie}
Taylor Apgar, Patrick Clary, Kevin Green, Alan Fern, and Jonathan Hurst.
\newblock Fast online trajectory optimization for the bipedal robot cassie.
\newblock In \emph{Robotics: Science and Systems}, 2018.

\bibitem[Berseth et~al.(2018)Berseth, Xie, Cernek, and van~de
  Panne]{2018-ICLR-distill}
Glen Berseth, Cheng Xie, Paul Cernek, and Michiel van~de Panne.
\newblock Progressive reinforcement learning with distillation for
  multi-skilled motion control.
\newblock In \emph{Proc. International Conference on Learning Representations},
  2018.

\bibitem[{Chen} et~al.(2018){Chen}, {Tangkaratt}, {Lin}, and
  {Sugiyama}]{2018-arxiv-Q_learning_demonstration}
Si-An {Chen}, Voot {Tangkaratt}, Hsuan-Tien {Lin}, and Masashi {Sugiyama}.
\newblock {Active Deep Q-learning with Demonstration}.
\newblock \emph{arXiv e-prints}, art. arXiv:1812.02632, December 2018.

\bibitem[Chen et~al.(2018)Chen, Ghadirzadeh, Folkesson, and
  Jensfelt]{2018-iros-acquire_nav}
Xi~Chen, Ali Ghadirzadeh, John Folkesson, and Patric Jensfelt.
\newblock Deep reinforcement learning to acquire navigation skills for
  wheel-legged robots in complex environments.
\newblock \emph{CoRR}, abs/1804.10500, 2018.
\newblock URL \url{http://arxiv.org/abs/1804.10500}.

\bibitem[Clavera et~al.(2018)Clavera, Nagabandi, Fearing, Abbeel, Levine, and
  Finn]{2019-iclr-learning_to_adapt}
Ignasi Clavera, Anusha Nagabandi, Ronald~S. Fearing, Pieter Abbeel, Sergey
  Levine, and Chelsea Finn.
\newblock Learning to adapt: Meta-learning for model-based control.
\newblock \emph{CoRR}, abs/1803.11347, 2018.
\newblock URL \url{http://arxiv.org/abs/1803.11347}.

\bibitem[Da et~al.(2017)Da, Hartley, and Grizzle]{2017-icra-supervised_atrias}
X.~Da, R.~Hartley, and J.~W. Grizzle.
\newblock Supervised learning for stabilizing underactuated bipedal robot
  locomotion, with outdoor experiments on the wave field.
\newblock In \emph{2017 IEEE International Conference on Robotics and
  Automation (ICRA)}, pages 3476--3483, May 2017.
\newblock \doi{10.1109/ICRA.2017.7989397}.

\bibitem[Gong et~al.(2018)Gong, Hartley, Da, Hereid, Harib, Huang, and
  Grizzle]{2018-arxiv-cassie}
Yukai Gong, Ross Hartley, Xingye Da, Ayonga Hereid, Omar Harib, Jiunn{-}Kai
  Huang, and Jessy~W. Grizzle.
\newblock Feedback control of a cassie bipedal robot: Walking, standing, and
  riding a segway.
\newblock \emph{CoRR}, abs/1809.07279, 2018.
\newblock URL \url{http://arxiv.org/abs/1809.07279}.

\bibitem[Heess et~al.(2017)Heess, TB, Sriram, Lemmon, Merel, Wayne, Tassa,
  Erez, Wang, Eslami, Riedmiller, and Silver]{2017-arxiv-parkour}
Nicolas Heess, Dhruva TB, Srinivasan Sriram, Jay Lemmon, Josh Merel, Greg
  Wayne, Yuval Tassa, Tom Erez, Ziyu Wang, S.~M.~Ali Eslami, Martin~A.
  Riedmiller, and David Silver.
\newblock Emergence of locomotion behaviours in rich environments.
\newblock \emph{CoRR}, abs/1707.02286, 2017.
\newblock URL \url{http://arxiv.org/abs/1707.02286}.

\bibitem[Henderson et~al.(2018)Henderson, Islam, Bachman, Pineau, Precup, and
  Meger]{2018-AAAI-DRL_matters}
Peter Henderson, Riashat Islam, Philip Bachman, Joelle Pineau, Doina Precup,
  and David Meger.
\newblock Deep reinforcement learning that matters.
\newblock In \emph{AAAI}, 2018.

\bibitem[Hereid et~al.(2018{\natexlab{a}})Hereid, Hubicki, Cousineau, and
  Ames]{2018-TRO-gait_opt}
A.~Hereid, C.~M. Hubicki, E.~A. Cousineau, and A.~D. Ames.
\newblock Dynamic humanoid locomotion: A scalable formulation for hzd gait
  optimization.
\newblock \emph{IEEE Transactions on Robotics}, 34\penalty0 (2):\penalty0
  370--387, April 2018{\natexlab{a}}.
\newblock ISSN 1552-3098.
\newblock \doi{10.1109/TRO.2017.2783371}.

\bibitem[Hereid et~al.(2018{\natexlab{b}})Hereid, Harib, Hartley, Gong, and
  Grizzle]{2018-arxiv-cfrost}
Ayonga Hereid, Omar Harib, Ross Hartley, Yukai Gong, and Jessy~W. Grizzle.
\newblock Rapid bipedal gait design using {C-FROST} with illustration on a
  cassie-series robot.
\newblock \emph{CoRR}, abs/1807.06614, 2018{\natexlab{b}}.
\newblock URL \url{http://arxiv.org/abs/1807.06614}.

\bibitem[Hirai et~al.(1998)Hirai, Hirose, Haikawa, and
  Takenaka]{1998-icra-asimo}
K.~Hirai, M.~Hirose, Y.~Haikawa, and T.~Takenaka.
\newblock The development of honda humanoid robot.
\newblock In \emph{Proceedings. 1998 IEEE International Conference on Robotics
  and Automation (Cat. No.98CH36146)}, volume~2, pages 1321--1326 vol.2, May
  1998.
\newblock \doi{10.1109/ROBOT.1998.677288}.

\bibitem[{Ho} and {Ermon}(2016)]{2016-arxiv-gail}
Jonathan {Ho} and Stefano {Ermon}.
\newblock {Generative Adversarial Imitation Learning}.
\newblock \emph{arXiv e-prints}, art. arXiv:1606.03476, June 2016.

\bibitem[Hwangbo et~al.(2019)Hwangbo, Lee, Dosovitskiy, Bellicoso, Tsounis,
  Koltun, and Hutter]{2019-scirobotics-anymal_learning}
Jemin Hwangbo, Joonho Lee, Alexey Dosovitskiy, Dario Bellicoso, Vassilios
  Tsounis, Vladlen Koltun, and Marco Hutter.
\newblock Learning agile and dynamic motor skills for legged robots.
\newblock \emph{Science Robotics}, 4\penalty0 (26), 2019.
\newblock \doi{10.1126/scirobotics.aau5872}.
\newblock URL \url{http://robotics.sciencemag.org/content/4/26/eaau5872}.

\bibitem[Kajita et~al.(2001)Kajita, Kanehiro, Kaneko, Yokoi, and
  Hirukawa]{2001-iros-LIP}
S.~Kajita, F.~Kanehiro, K.~Kaneko, K.~Yokoi, and H.~Hirukawa.
\newblock The 3d linear inverted pendulum mode: a simple modeling for a biped
  walking pattern generation.
\newblock In \emph{Proceedings 2001 IEEE/RSJ International Conference on
  Intelligent Robots and Systems. Expanding the Societal Role of Robotics in
  the the Next Millennium (Cat. No.01CH37180)}, volume~1, pages 239--246 vol.1,
  Oct 2001.
\newblock \doi{10.1109/IROS.2001.973365}.

\bibitem[Kingma and Ba(2014)]{2014-arxiv-adam}
Diederik~P. Kingma and Jimmy Ba.
\newblock Adam: {A} method for stochastic optimization.
\newblock \emph{CoRR}, abs/1412.6980, 2014.
\newblock URL \url{http://arxiv.org/abs/1412.6980}.

\bibitem[Laskey et~al.(2017)Laskey, Lee, Fox, Dragan, and
  Goldberg]{2017-corl-dart}
Michael Laskey, Jonathan Lee, Roy Fox, Anca~D. Dragan, and Kenneth~Y. Goldberg.
\newblock Dart: Noise injection for robust imitation learning.
\newblock In \emph{CoRL}, 2017.

\bibitem[Levine and Koltun(2013)]{2013-icml-gps}
Sergey Levine and Vladlen Koltun.
\newblock Guided policy search.
\newblock In Sanjoy Dasgupta and David McAllester, editors, \emph{Proceedings
  of the 30th International Conference on Machine Learning}, volume~28 of
  \emph{Proceedings of Machine Learning Research}, pages 1--9, Atlanta,
  Georgia, USA, 17--19 Jun 2013. PMLR.
\newblock URL \url{http://proceedings.mlr.press/v28/levine13.html}.

\bibitem[Li et~al.(2018)Li, Rai, Geyer, and Atkeson]{2018-arxiv-rl_on_atrias}
Tianyu Li, Akshara Rai, Hartmut Geyer, and Christopher~G. Atkeson.
\newblock Using deep reinforcement learning to learn high-level policies on the
  {ATRIAS} biped.
\newblock \emph{CoRR}, abs/1809.10811, 2018.
\newblock URL \url{http://arxiv.org/abs/1809.10811}.

\bibitem[Liu et~al.(2013)Liu, Atkeson, and
  Su]{2013-robotica-trajectory_library}
Chenggang Liu, Christopher~G. Atkeson, and Jianbo Su.
\newblock Biped walking control using a trajectory library.
\newblock \emph{Robotica}, 31\penalty0 (2):\penalty0 311–322, 2013.
\newblock \doi{10.1017/S0263574712000203}.

\bibitem[Merel et~al.(2018)Merel, Hasenclever, Galashov, Ahuja, Pham, Wayne,
  Teh, and Heess]{2018-arxiv-neural_primitive}
Josh Merel, Leonard Hasenclever, Alexandre Galashov, Arun Ahuja, Vu~Pham, Greg
  Wayne, Yee~Whye Teh, and Nicolas Heess.
\newblock Neural probabilistic motor primitives for humanoid control.
\newblock \emph{CoRR}, abs/1811.11711, 2018.
\newblock URL \url{http://arxiv.org/abs/1811.11711}.

\bibitem[Nakanishi et~al.(2004)Nakanishi, Morimoto, Endo, Cheng, Schaal, and
  Kawato]{2004-RAS-lfd_biped}
Jun Nakanishi, Jun Morimoto, Gen Endo, Gordon Cheng, Stefan Schaal, and Mitsuo
  Kawato.
\newblock Learning from demonstration and adaptation of biped locomotion.
\newblock \emph{Robotics and Autonomous Systems}, 47\penalty0 (2):\penalty0 79
  -- 91, 2004.
\newblock ISSN 0921-8890.
\newblock \doi{https://doi.org/10.1016/j.robot.2004.03.003}.
\newblock URL
  \url{http://www.sciencedirect.com/science/article/pii/S0921889004000399}.
\newblock Robot Learning from Demonstration.

\bibitem[Parisotto et~al.(2015)Parisotto, Ba, and
  Salakhutdinov]{2015-arxiv-actor_mimic}
Emilio Parisotto, Lei~Jimmy Ba, and Ruslan Salakhutdinov.
\newblock Actor-mimic: Deep multitask and transfer reinforcement learning.
\newblock \emph{CoRR}, abs/1511.06342, 2015.
\newblock URL \url{http://arxiv.org/abs/1511.06342}.

\bibitem[Peng et~al.(2017)Peng, Berseth, Yin, and van~de
  Panne]{2017-TOG-deepLoco}
Xue~Bin Peng, Glen Berseth, KangKang Yin, and Michiel van~de Panne.
\newblock Deeploco: Dynamic locomotion skills using hierarchical deep
  reinforcement learning.
\newblock \emph{ACM Transactions on Graphics (Proc. SIGGRAPH 2017)},
  36\penalty0 (4), 2017.

\bibitem[Peng et~al.(2018)Peng, Abbeel, Levine, and van~de
  Panne]{2018-TOG-deepMimic}
Xue~Bin Peng, Pieter Abbeel, Sergey Levine, and Michiel van~de Panne.
\newblock Deepmimic: Example-guided deep reinforcement learning of
  physics-based character skills.
\newblock \emph{ACM Transactions on Graphics (Proc. SIGGRAPH 2018)},
  37\penalty0 (4), 2018.

\bibitem[Pomerleau(1988)]{1988-nips-ALVINNAA}
Dean Pomerleau.
\newblock Alvinn: An autonomous land vehicle in a neural network.
\newblock In \emph{NIPS}, 1988.

\bibitem[Posa et~al.(2016)Posa, Kuindersma, and
  Tedrake]{2016-icra-constrained_systems}
Michael Posa, Scott Kuindersma, and Russ Tedrake.
\newblock Optimization and stabilization of trajectories for constrained
  dynamical systems.
\newblock In \emph{Proceedings of the International Conference on Robotics and
  Automation}, 2016.

\bibitem[Ross and Bagnell(2010)]{2010-ICAIS-covariate_shift}
Stephane Ross and Drew Bagnell.
\newblock Efficient reductions for imitation learning.
\newblock In Yee~Whye Teh and Mike Titterington, editors, \emph{Proceedings of
  the Thirteenth International Conference on Artificial Intelligence and
  Statistics}, volume~9 of \emph{Proceedings of Machine Learning Research},
  pages 661--668, Chia Laguna Resort, Sardinia, Italy, 13--15 May 2010. PMLR.
\newblock URL \url{http://proceedings.mlr.press/v9/ross10a.html}.

\bibitem[Ross et~al.(2011)Ross, Gordon, and Bagnell]{2011-AISTATS-dagger}
St{\'e}phane Ross, Geoffrey~J. Gordon, and J.~Andrew Bagnell.
\newblock A reduction of imitation learning and structured prediction to
  no-regret online learning.
\newblock In \emph{AISTATS}, 2011.

\bibitem[Rusu et~al.(2015)Rusu, Colmenarejo, G{\"{u}}l{\c{c}}ehre, Desjardins,
  Kirkpatrick, Pascanu, Mnih, Kavukcuoglu, and Hadsell]{2015-arxiv-distill}
Andrei~A. Rusu, Sergio~Gomez Colmenarejo, {\c{C}}aglar G{\"{u}}l{\c{c}}ehre,
  Guillaume Desjardins, James Kirkpatrick, Razvan Pascanu, Volodymyr Mnih,
  Koray Kavukcuoglu, and Raia Hadsell.
\newblock Policy distillation.
\newblock \emph{CoRR}, abs/1511.06295, 2015.
\newblock URL \url{http://arxiv.org/abs/1511.06295}.

\bibitem[Schaal et~al.(2003)Schaal, Peters, Nakanishi, and
  Ijspeert]{2003-iros-DMP}
S.~Schaal, J.~Peters, J.~Nakanishi, and A.~Ijspeert.
\newblock Control, planning, learning, and imitation with dynamic movement
  primitives.
\newblock In \emph{IROS 2003}, pages 1--21. Max-Planck-Gesellschaft, October
  2003.

\bibitem[Schuitema et~al.(2010)Schuitema, Wisse, Ramakers, and
  Jonker]{2010-iros-2dleo}
E.~Schuitema, M.~Wisse, T.~Ramakers, and P.~Jonker.
\newblock The design of leo: A 2d bipedal walking robot for online autonomous
  reinforcement learning.
\newblock In \emph{2010 IEEE/RSJ International Conference on Intelligent Robots
  and Systems}, pages 3238--3243, Oct 2010.
\newblock \doi{10.1109/IROS.2010.5650765}.

\bibitem[Schulman et~al.(2017)Schulman, Wolski, Dhariwal, Radford, and
  Klimov]{2017-arxiv-ppo}
John Schulman, Filip Wolski, Prafulla Dhariwal, Alec Radford, and Oleg Klimov.
\newblock Proximal policy optimization algorithms.
\newblock \emph{CoRR}, abs/1707.06347, 2017.
\newblock URL \url{http://arxiv.org/abs/1707.06347}.

\bibitem[Sutton et~al.(1999)Sutton, McAllester, Singh, and
  Mansour]{1999-nips-policy_gradient}
Richard~S. Sutton, David~A. McAllester, Satinder~P. Singh, and Yishay Mansour.
\newblock Policy gradient methods for reinforcement learning with function
  approximation.
\newblock In \emph{NIPS}, 1999.

\bibitem[Tan et~al.(2018)Tan, Zhang, Coumans, Iscen, Bai, Hafner, Bohez, and
  Vanhoucke]{2018-rss-sim_to_real}
Jie Tan, Tingnan Zhang, Erwin Coumans, Atil Iscen, Yunfei Bai, Danijar Hafner,
  Steven Bohez, and Vincent Vanhoucke.
\newblock Sim-to-real: Learning agile locomotion for quadruped robots.
\newblock \emph{CoRR}, abs/1804.10332, 2018.

\bibitem[Tang and Hauser(2017)]{2017-iros-data_driven_control}
G.~Tang and K.~Hauser.
\newblock A data-driven indirect method for nonlinear optimal control.
\newblock In \emph{2017 IEEE/RSJ International Conference on Intelligent Robots
  and Systems (IROS)}, pages 4854--4861, Sept 2017.
\newblock \doi{10.1109/IROS.2017.8206362}.

\bibitem[Tang et~al.(2018)Tang, Sun, and
  Hauser]{2018-iros-learning_trajectories}
Gao Tang, Weidong Sun, and Kris Hauser.
\newblock Learning trajectories for real-time optimal control of quadrotors.
\newblock In \emph{2018 IEEE/RSJ International Conference on Intelligent Robots
  and Systems (IROS)}, 2018.

\bibitem[Tedrake et~al.(2004)Tedrake, Zhang, and
  Seung]{2004-iros-policy_gradient_biped}
Russ Tedrake, Teresa~Weirui Zhang, and H.~Sebastian Seung.
\newblock Stochastic policy gradient reinforcement learning on a simple 3d
  biped.
\newblock \emph{2004 IEEE/RSJ International Conference on Intelligent Robots
  and Systems (IROS) (IEEE Cat. No.04CH37566)}, 3:\penalty0 2849--2854 vol.3,
  2004.

\bibitem[Tedrake et~al.(2015)Tedrake, Kuindersma, Deits, and
  Miura]{2015-humanoids-ZMP_atlas}
Russ Tedrake, Scott Kuindersma, Robin Deits, and Kanako Miura.
\newblock A closed-form solution for real-time zmp gait generation and feedback
  stabilization.
\newblock In \emph{IEEE-RAS International Conference on Humanoid Robots},
  Seoul, Korea, 2015.

\bibitem[Todorov et~al.(2012)Todorov, Erez, and Tassa]{2012-iros-mujoco}
E.~Todorov, T.~Erez, and Y.~Tassa.
\newblock Mujoco: A physics engine for model-based control.
\newblock In \emph{2012 IEEE/RSJ International Conference on Intelligent Robots
  and Systems}, pages 5026--5033, Oct 2012.
\newblock \doi{10.1109/IROS.2012.6386109}.

\bibitem[Xie et~al.(2018)Xie, Berseth, Clary, Hurst, and van~de
  Panne]{2018-IROS-cassie}
Zhaoming Xie, Glen Berseth, Patrick Clary, Jonathan Hurst, and Michiel van~de
  Panne.
\newblock Feedback control for cassie with deep reinforcement learning.
\newblock In \emph{Proc. IEEE/RSJ Intl Conf on Intelligent Robots and Systems
  (IROS 2018)}, 2018.

\bibitem[Xiong and Ames(2018)]{2018-humanoids-coupling}
Xiaobin Xiong and Aaron~D. Ames.
\newblock Coupling reduced order models via feedback control for 3d
  underactuated bipedal robotic walking.
\newblock \emph{IEEE-RAS 18th International Conference on Humanoid Robots
  (Humanoids)}, 2018.
\newblock URL \url{http://ames.caltech.edu/xiong2018coupling.pdf}.

\bibitem[Yu et~al.(2018)Yu, Turk, and Liu]{2018-SIGGRAPH-symmetry}
Wenhao Yu, Greg Turk, and C.~Karen Liu.
\newblock Learning symmetric and low-energy locomotion.
\newblock \emph{ACM Transactions on Graphics (Proc. SIGGRAPH 2018 - to
  appear)}, 37\penalty0 (4), 2018.

\end{thebibliography}
